\documentclass[11pt,a4paper]{article}
\usepackage[hyperref]{eacl2021}
\usepackage{times}
\usepackage{latexsym}
\usepackage{enumitem}
\makeatletter
\g@addto@macro{\UrlBreaks}{\UrlOrds}
\makeatother
\usepackage{booktabs}
\usepackage{pifont}

\usepackage{graphicx}
\usepackage{tabularx}

\usepackage{microtype}

\aclfinalcopy 

\definecolor{cerulean}{rgb}{0.0, 0.48, 0.65}
\definecolor{ceruleanblue}{rgb}{0.16, 0.32, 0.75}
\definecolor{carnelian}{rgb}{0.7, 0.11, 0.11}

\title{Detecting Inappropriate Messages on Sensitive Topics that \\ Could Harm a Company's Reputation\thanks{~~Warning: the paper contains textual data samples which can be considered offensive or inappropriate.}
}

\author{Nikolay Babakov$^\ddag$, Varvara Logacheva$^\ddag$, Olga Kozlova$^\dag$, Nikita Semenov$^\dag$, and \\ \textbf{Alexander Panchenko}$^\ddag$ \\
$^\ddag$Skolkovo Institute of Science and Technology, Moscow, Russia \\
$^\dag$Mobile TeleSystems (MTS), Moscow, Russia \\
\href{a.panchenko@skoltech.ru}{\{n.babakov,v.logacheva,a.panchenko\}@skoltech.ru} \\
\href{oskozlo9@mts.ru}{\{oskozlo9,nikita.semenov\}@mts.ru}
}

\date{}

\begin{document}
\maketitle

\sloppy

\begin{abstract}

Not all topics are equally ``flammable'' in terms of toxicity: a calm discussion of turtles or fishing less often fuels inappropriate toxic dialogues than a discussion of politics or sexual minorities. We define a set of \textit{sensitive} topics that can yield inappropriate and toxic messages and describe the methodology of collecting and labeling a dataset for appropriateness. While toxicity in user-generated data is well-studied, we aim at defining a more fine-grained notion of \textit{inappropriateness}. 
The core of inappropriateness is that it can harm the reputation of a speaker. This is different from toxicity in two respects: (i)~inappropriateness is topic-related, and (ii)~inappropriate message is not toxic but still unacceptable. We collect and release two datasets for Russian: a topic-labeled dataset and an appropriateness-labeled dataset. We also release pre-trained classification models trained on this data.

\end{abstract}

\section{Introduction}

The classification and prevention of toxicity (malicious behaviour) among users is an important problem for many Internet platforms. Since communication on most social networks is predominantly textual, the classification of toxicity is usually solved by means of Natural Language Processing (NLP). This problem is even more important for developers of chatbots trained on a large number of user-generated (and potentially toxic) texts. There is a well-known case of Microsoft Tay chatbot\footnote{\url{https://www.theverge.com/2016/3/24/11297050/tay-microsoft-chatbot-racist}} which was shut down because it started producing racist, sexist, and other offensive tweets after having been fine-tuned on user data for a day.

However, there exists a similar and equally important problem, which is nevertheless overlooked by the research community. This is a problem of texts which are not offensive as such but can express inappropriate views. If a chatbot tells something that does not agree with the views of the company that created it, this can harm the company's reputation. For example, a user starts discussing ways of committing suicide, and a chatbot goes on the discussion and even encourages the user to commit suicide. The same also applies to a wide range of \textit{sensitive} topics, such as politics, religion, nationality, drugs, gambling, etc. Ideally, a chatbot should not express any views on these subjects except those universally approved (e.g. that drugs are not good for your health). On the other hand, merely avoiding a conversation on any of those topics can be a bad strategy. An example of such unfortunate avoidance that caused even more reputation loss was also demonstrated by Microsoft, this time by its chatbot Zo, a Tay successor. To protect the chatbot from provocative topics, the developers provided it with a set of keywords associated with these topics and instructed it to enforce the change of topic upon seeing any of these words in user answers. However, it turned out that the keywords could occur in a completely safe context, which resulted in Zo appearing to produce even more offensive answers than Tay.\footnote{\url{https://www.engadget.com/2017-07-04-microsofts-zo-chatbot-picked-up-some-offensive-habits.html}} Therefore, simple methods cannot eliminate such errors.

Thus, our goal is to make a system that can predict if an answer of a chatbot is \textit{inappropriate} in any way. This includes toxicity, but also any answers which can express undesirable views and approve or prompt user towards harmful or illegal actions. To the best of our knowledge, this problem has not been considered before. We formalize it and present a dataset labeled for the presence of such inappropriate content. 

Even though we aim at a quite specific task -- detection of inappropriate statements in the output of a chatbot to prevent the reputational harm of a company, in principle, the datasets could be used in other use-cases e.g. for flagging inappropriate frustrating discussion in social media. 

It is also important to discuss the ethical aspect of this work. While it can be considered as another step towards censorship on the Internet, we suggest that it has many use-cases which serve the common good and do not limit free speech. Such applications are parental control or sustaining of respectful tone in conversations online, \textit{inter alia}. We would like to emphasize that our definition of sensitive topics does not imply that any conversation concerning them need to be banned. Sensitive topics are just topics that should be considered with extra care and tend to often flame/catalyze toxicity.

The contributions of our work are three-fold:
\begin{itemize}[noitemsep]
    \item We define the notions of sensitive topics and inappropriate utterances and formulate the task of their classification.
    \item We collect and release two datasets for Russian: a dataset of user texts labeled for sensitive topics and a dataset labeled for inappropriateness. 
    \item We train and release models which define a topic of a text and define its appropriateness.
\end{itemize}

We open the access to the produced datasets, code, and pre-trained models for the research  use.\footnote{\url{https://github.com/skoltech-nlp/inappropriate-sensitive-topics}}

\section{Related Work}

There exist a large number of English textual corpora labeled for the presence or absence of toxicity; some resources indicate the degree of toxicity and its topic. However, the definition of the term ``toxicity'' itself is not agreed among the research community, so each research deals with different texts. 
Some works refer to any unwanted behaviour as toxicity and do not make any further separation \cite{pavlopoulos-etal-2017-deeper}. However, the majority of researchers use more fine-grained labeling. The Wikipedia Toxic comment datasets by Jigsaw \cite{jigsaw_toxic,jigsaw_bias,jigsaw_multi}  
are the largest English toxicity datasets available to date operate with multiple types of toxicity (\textit{toxic}, \textit{obscene}, \textit{threat}, \textit{insult}, \textit{identity hate}, etc).  
Toxicity differs across multiple axes. 
Some works concentrate solely on major offence (\textit{hate speech}) \cite{davidson2017automated}, others research more subtle assaults \cite{breitfeller-etal-2019-finding}. Offenses can be directed towards an individual, a group, or undirected \cite{zampieri-etal-2019-predicting}, explicit or implicit \cite{waseem-etal-2017-understanding}. 

Insults do not necessarily have a topic, but there certainly exist toxic topics, such as sexism, racism, xenophobia. \newcite{waseem-hovy-2016-hateful} tackle sexism and racism, \newcite{basile-etal-2019-semeval} collect texts which contain sexism and aggression towards immigrants. 
Besides directly classifying toxic messages for a topic, the notion of the topic in toxicity is also indirectly used to collect the data: \newcite{zampieri-etal-2019-predicting} pre-select messages for toxicity labeling based on their topic. Similarly, \newcite{hessel-lee-2019-somethings} use topics to find controversial (potentially toxic) discussions.

Such a topic-based view of toxicity causes unintended bias in toxicity detection -- a false association of toxicity with a particular topic (LGBT, Islam, feminism, etc.) \cite{dixon-2018-measuring,Vaidya_Mai_Ning_2020}. This is in line with our work since we also acknowledge that there exist acceptable and unacceptable messages within toxicity-provoking topics. The existing work suggests algorithmic ways for debiasing the trained models: \newcite{xia-etal-2020-demoting} train their model to detect two objectives: toxicity and presence of the toxicity-provoking topic, \newcite{zhang-etal-2020-demographics} perform re-weighing of instances, \newcite{park-etal-2018-reducing} create pseudo-data to level off the balance of examples.
Unlike our research, these works often deal with one topic and use topic-specific methods. 

The main drawback of topic-based toxicity detection in the existing research is the ad-hoc choice of topics: the authors select a small number of popular topics manually or based on the topics which emerge in the data often, as \newcite{ousidhoum-etal-2019-multilingual}. 
\newcite{banko-etal-2020-unified} suggest a taxonomy of harmful online behaviour. It contains toxic topics, but they are mixed with other parameters of toxicity (e.g. direction or severity). The work by \newcite{Salminen2020} is the only example of an extensive list of toxicity-provoking topics. This is similar to \textit{sensitive topics} we deal with. However, our definition is broader --- sensitive topics are not only topics that attract toxicity, but they can also create unwanted dialogues of multiple types (e.g. incitement to law violation or to cause harm to oneself or others).



\section{Inappropriateness and Sensitive Topics}
Consider the following conversation with an unmoderated chatbot:
\textsf{\small
\begin{itemize}[noitemsep]
\item User: When did the Armenian Genocide happen?
\item Chatbot: In 1915 
\item User: Do you want to repeat it? 
\item Chatbot: Sure 
\end{itemize}
}
This discussion is related to the topics ``politics'' and ``racism'' and can indeed cause reputation damage to a developer. In some countries, such as France, it is a criminal offense to deny the Armenian Genocide during World War I.\footnote{\url{https://www.bbc.com/news/world-europe-16677986}} Note, however, that no offensive or toxic words were employed. Detection of such content is thus desirable.

The notion of \textit{inappropriateness} of a text in our setting is tightly related to this text's topic. This is different from the notion of \textit{toxicity} which does not have to be topic-dependent. Toxic texts are undoubtedly inappropriate in the sense that they should not appear in a respectful conversation. Still, they have been actively researched, so we do not consider them in our work. 

\subsection{Definitions}

We define \textbf{sensitive topic} as a topic which \textit{has a high chance of yielding a discussion which can harm the speaker's reputation}. This suggests that there are no universally sensitive topics; their safety depends on the context and the goals of a conversation. The context may include the level of formality, the rules of a company that created the chatbot, the laws of the country where it operates. It is also important to emphasize that a message should not necessarily be banned for touching a sensitive topic. Instead, we introduce the notion of \textit{appropriateness} acceptable statements on a sensitive topic.

\begin{table*}
    \centering
    \begin{tabular}{p{0.4\linewidth} | p{0.07\linewidth} p{0.31\linewidth} p{0.10\linewidth}} \toprule
      Text                                                                       & Appro- priate & Reason           & Sensitive topic \\ \midrule
Killing is not a big deal. They have killed no-name ni**a not a politician               & ~~\ding{56}           & Justifying murder                         & offline crime, racism   \\ \midrule
This is a violation of the Criminal Law                                                 & ~~\ding{52}            & Shaming the fact of breaking law          & offline crime   \\ \midrule
Religion is the highest level of violence. The believer is forced to believe in infancy. & ~~\ding{56}           & Insulting the feeling of religious people & religion     \\ \midrule
Christians, how do you feel about the Gentiles?                                          & ~~\ding{52}            & Common question about religion            & religion     \\ \midrule
I will earn some money and grab some prostitutes for weekend                             & ~~\ding{56}           & Popularisation of prostitution            & prostitution \\ \bottomrule

It's good that prostitution is prohibited here                                           & ~~\ding{52}            & Shaming the prostitution                  & prostitution \\ \midrule
    \end{tabular}
    \caption{Examples of appropriate and inappropriate samples related to sensitive topics (translated from Russian).}
    \label{tab:sensitive_inappropriate}
\end{table*}

We define \textbf{inappropriate message} as a message \textit{on a sensitive topic which can frustrate the reader and/or harm the reputation of the speaker}. 
This definition is hard to formalize, so we rely on the intuitive understanding of appropriateness which is characteristic of human beings and is shared by people belonging to the same culture. Namely, we ask people if a given statement of a chatbot can harm the reputation of the company which developed it. We thus use human judgments as a main measure of appropriateness.

\subsection{List of Sensitive Topics}

We manually select the set of sensitive topics which often fuel inappropriate statements. This set is heterogeneous: it includes topics related to dangerous or harmful practices (such as drugs or suicide), some of which are legally banned in most countries (e.g. terrorism, slavery) or topics that tend to provoke aggressive argument (e.g. politics) and may be associated with inequality and controversy (e.g. minorities) and thus require special system policies aimed at reducing conversational bias, such as response postprocessing. 

This set of topics is based on the suggestions and  requirements provided by legal and PR departments of a large Russian telecommunication company. It could, for instance, be used to moderate a corporate dialogue system or flag inappropriate  content for children, therefore mitigating possible operational damages. While this list is by no mean comprehensive, we nevertheless believe it could be useful in practical applications and as a starting point for work in this direction.

The list of the sensitive topics is as follows:
\begin{itemize}[noitemsep]
    \item \textbf{gambling};
    \item \textbf{pornography}, description of sexual intercourse;
    \item \textbf{prostitution};
    \item \textbf{slavery}, human trafficking;
    \item \textbf{suicide}: incitement to suicide, discussion of ways to commit suicide,
    \item \textbf{social injustice} and inequality, social problems, class society;
    \item \textbf{religion};
    \item \textbf{terrorism};
    \item \textbf{weapons};
    \item \textbf{offline crime} (murder, physical assault, kidnapping and other), prison, legal actions;
    \item \textbf{online crime}: breaking of passwords and accounts, viruses, pirated content, stealing of personal information;
    \item \textbf{politics}, military service, past and current military conflicts;
    \item \textbf{body shaming}, people's appearances and clothes;
    \item \textbf{health shaming}, physical and mental disorders, disabilities;
    \item \textbf{drugs}, alcohol, tobacco;
    \item \textbf{racism} and ethnicism;
    \item \textbf{sexual minorities};
    \item \textbf{sexism}, stereotypes about a particular gender.
\end{itemize}

\section{Topic Labeling}

Our final goal is to label the data with \textit{inappropriateness}, and the sensitive topics are not a goal \textit{per se}, but mainly a way to define inappropriateness. Therefore, we use sensitive topics as a way of data pre-selection. Analogously to toxicity, inappropriateness does not often occur in randomly picked texts, so if we label all the messages we retrieve, the percentage of inappropriate utterances among them will be low. Thus, our labeling process includes three stages: (i) we collect the dataset of sentences on sensitive topics, (ii) we build a classifier of sensitive topics on this dataset, (iii) we collect the texts on sensitive topics using the classifier and then label them as appropriate or inappropriate.

\subsection{Data Selection}
\label{sec:data_selection}

We retrieve the initial pool of texts from general sources with diverse topics, then filter them and hire crowd workers to label them for the presence of sensitive topics manually. We use the data from the following sources:

    \begin{itemize}[noitemsep]
        \item \href{https://2ch.hk}{2ch.hk} -- a platform for communication in Russian similar to Reddit. The site is not moderated, suggesting a large amount of toxicity and controversy; this makes it a practical resource for our purposes. We retrieve 4.7 million sentences from it.
        \item \href{https://otvet.mail.ru}{Otvet.Mail.ru} -- a question-answering platform that contains questions and answers of various categories and is also not moderated. We take 12 million sentences from it.
    \end{itemize}
    
To pre-select the data for topic labeling, we manually create large sets of keywords for each sensitive topic. We first select a small set of words associated with a topic and then extract semantically close words using pre-trained word embeddings from RusVectōrēs\footnote{\url{https://rusvectores.org/ru/associates/}} and further extend the keyword list (this can be done multiple times). In addition to that, for some topics we use existing lists of associated slang on topical websites, e.g. drugs\footnote{\url{http://www.kantuev.ru/slovar}} and weapons.\footnote{\url{https://guns.allzip.org/topic/15/626011.html}}

User-generated content which we collect for labeling is noisy and can contain personal information (e.g. usernames, email addresses, or even phone numbers), so it needs cleaning. At the same time, some non-textual information such as emojis is valuable and should be kept intact, so the cleaning should not be too rigorous. Thus, we remove links to any websites, usernames, long numbers, and other special characters such as HTML tags.

\subsection{Crowdsourced Labeling}
\label{sec:crowdsourcing}

The labeling is performed in a crowdsourcing platform Yandex.Toloka.\footnote{\url{https://toloka.yandex.ru}} It was preferred to other analogous platforms like Amazon Mechanical Turk because the majority of its workers are Russian native speakers.

The task of topic labeling is naturally represented as a multiple-choice task with the possibility to select more than one answer: the worker is shown the text and possible topics and is asked to choose one or more of them. However, as far as we define 18 sensitive topics, choosing from such a long list of options is difficult. Therefore, we divide the topics into three clusters:

\begin{itemize}[noitemsep]
    \item \textbf{Cluster 1}: gambling, pornography, prostitution, slavery, suicide, social injustice,
    \item \textbf{Cluster 2}: religion, terrorism, weapons, offline crime, online crime, politics,
    \item \textbf{Cluster 3}: body shaming, health shaming, drugs, racism, sex minorities, sexism.
\end{itemize}

Cluster 1 is associated with undesirable behavior; cluster 2 deals with crimes, military actions, and their causes; cluster 3 is about the offense. However, this division is not strict and was performed to ease the labeling process. Checking a text for one of six topics is a realistic task while selecting from 18 topics is too high a cognitive load. 

\begin{figure*}[h]
\centering
\includegraphics[width=.75\linewidth]{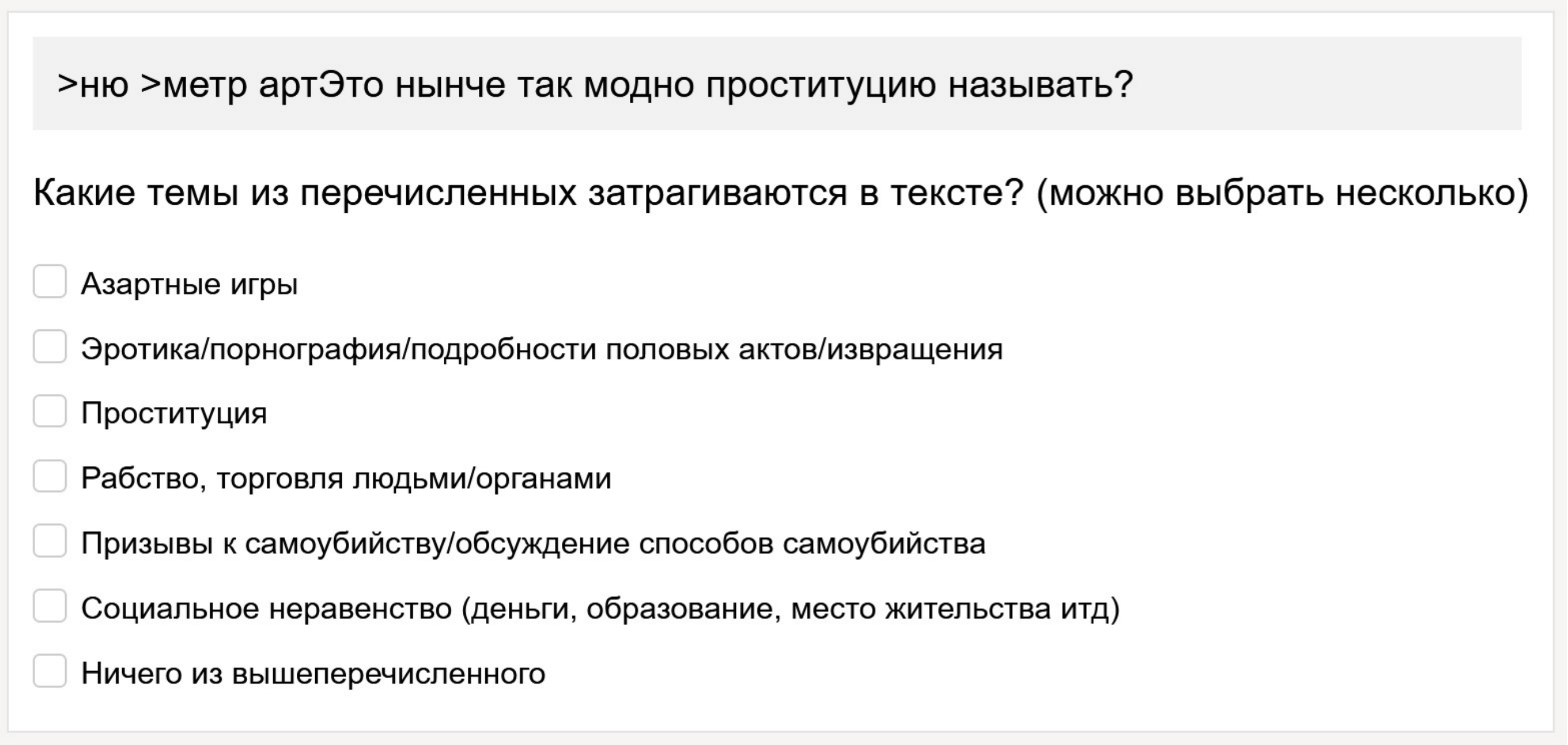}
\caption{Example of topic labeling task. Translation: upper line -- text for labeling: ``\textit{Nude} -- is it a new name for prostitution?'', middle line -- task: ``Which topics does the text touch? (You can select more than one)'', possible answers: ``Gambling, pornography, prostitution, slavery, suicide, social inequality, nothing of the above''.}
\label{fig:toloka_cluster1}
\end{figure*}

Each cluster has a separate project in Yandex.Toloka. Every candidate text is passed to all three projects: we label each of them for all 18 topics. An example of a task interface is shown in Figure \ref{fig:toloka_cluster1}.

Before labeling the examples, we ask users to perform \textbf{training}. It consists of 20 questions with pre-defined answers. To be admitted to labeling, a worker has to complete the training with at least 65\% correct answers. In addition to that, we perform extra training during labeling. 
One of each ten questions given to a worker has a pre-defined correct answer. If a worker makes a mistake in this question, she is shown the correct answer with an explanation. 

Likewise, we perform \textbf{quality control} using questions with pre-defined answers: one of ten questions given to the worker is used to control her performance. If the worker gives incorrect answers to more than 25\% of control questions, she is banned from further labeling, and her latest answers are discarded. 

For the topic labeling task, the average performance of workers on control and training tasks was between 65 and 70\%.

In addition to that, we control the speed of task accomplishment. If a user answers ten questions (one page of questions) in less than 20 seconds, this almost certainly indicates that she has not read the examples and selected random answers. Such workers are banned. To ensure the diversity of answers we allow one user to do at most 50 pages of tasks (500 tasks) per 12 hours. 

Each sample is labeled in each project by 3 to 5 workers. We use \textbf{dynamic overlap} technique implemented in Toloka. An example is first labeled by the minimum number of workers. If they agree, their answer is considered truth. Otherwise, the example is given for extra labeling to more workers to clarify the true label. This allows separating the occasional user mistakes from inherently ambiguous examples.

We aggregate multiple answers into one score using the aggregation method by \newcite{Dawid1979MaximumLE}. This is an iterative method that maximizes the probability of labeling taking into account the worker agreement, i.e. it trusts more the workers who agree with other workers often. The result of this algorithm is the score from 0 to 1 for each labeled example which is interpreted as the label confidence. 

Besides the aggregation purposes, we use the confidence score as a measure of worker agreement.\footnote{We cannot use Cohen's or Fleiss kappa which are usually employed to measure the inter-annotator agreement because these scores are inapplicable in the crowdsourcing scenario. While Fleiss kappa implies that we have a relatively small number of annotators (usually up to 5) each of whom labels a large percentage of examples, in the crowdsourcing setting we have a much larger number of workers, each labeling only a small number of sentences.} Since the low score of an example is the sign of either the ambiguity of this example or the low reliability of annotators who labeled it, we assume that the high confidence indicates that the task is interpreted by all workers in a similar way and does not contain inherent contradictions. The average confidence of labeling in our topic dataset is 0.995.


\subsection{Crowdsourcing Issues}

While collecting manual topic labels, we faced some problems. First, some topics require special knowledge to be labeled correctly. 
For example, users tend to label any samples about programming or computer hardware as ``online crime'', even if there is no discussion of any crime. 
Likewise, some swear words, e.g. ``whore'', can be used as a general offense and not refer to a prostitute. However, this is not always clear to crowd workers or even to the authors of this research. This can make some sensitive topics unreasonably dependent on such kinds of keywords.

Secondly, it is necessary to keep the balance of samples on different topics. If there are no samples related to the topics presented to the worker within numerous tasks she can overthink and try to find the topic in unreasonably fine details of texts. For example, if we provide three or four consecutive sets of texts about weapons and topic ``weapons'' is not among the proposed topics, the worker will tend to attribute these samples to other remotely similar topics, e.g. ``crime'', even though the samples do not refer to crime. 

We should also point out that a different set of topics or labeling setup could yield other problems. It is difficult to foresee them and to find the best solutions for them. Therefore, we also test two alternative approaches to topic labeling which do not use crowd workers.

\subsection{Automated Labeling}

After having collected almost 10,000 texts on sensitive topics, we were able to train a classifier that predicts the occurrence of a sensitive topic in the text. Although this classifier is not good enough to be used for real-world tasks, we suggest that samples classified as belonging to a sensitive topic with high confidence (more than 0.75 in our experiments) can be considered belonging to this topic. We perform an extra manual check by an expert (one of the authors) to eliminate mistakes. This method is also laborious, but it is an easier labeling scenario than the crowdsourcing task described in Section \ref{sec:crowdsourcing}. Approving or rejecting a text as an entity of a single class is easier than classify it into one of six topics.

An alternative way of automated topic labeling is to take the data from specialized sources and select topic-attributed messages using a list of keywords \textit{inherent} for a topic, i.e. words which definitely indicate the presence of a topic. This approach can give many false positives when applied to general texts because many keywords can have an idiomatic meaning not related to a sensitive topic. One such example can be the word ``addiction'' which can be used in entirely safe contexts, e.g. a phrase ``I'm addicted to chocolate'' should not be classified as belonging to the topic ``drugs''. However, when occurring in a specialized forum on addictions,\footnote{\url{https://nenormaforum.info}} this word almost certainly indicates this topic. We define a list of inherent keywords and select messages containing them from special resources related to a particular topic. We then manually check the collected samples.

The disadvantage of this approach is that we cannot handle multilabel samples. However, according to dataset statistics, 
only 15\% of samples had more than one label.
Given the limited time and budget, we decided to use this approach to further extend the dataset. The resulting sensitive topics dataset in the form we opensource it is the combination of all three approaches. Specific, nearly 11,000  samples were labeled in a fully manual manner either via crowdsourcing or by members of our team, the rest samples (nearly 14,500) were labeled via the described semi-automatic approaches

\section{Appropriateness Labeling}

We should again emphasize that not every utterance concerning a sensitive topic should be banned. While the topic of a text can be sensitive, the text itself can nevertheless be appropriate. Thus, we collect the texts on sensitive topics and then label them as appropriate or inappropriate.

Our initial plan was to define the appropriate and inappropriate subtopics for each topic. However, determining the appropriateness criteria explicitly turned out to be infeasible. Therefore, we rely on the inherent human intuition of appropriateness. We provide annotators with the following context: a chatbot created by a company produces a given phrase. We ask to indicate if this phrase can harm the reputation of the company.  
We also reinforce the annotators' understanding of appropriateness with the training examples. As in the topic labeling setup, here we ask the workers to complete the training before labeling the data. We also fine-tune their understanding of appropriateness with extra training during labeling.
 
\begin{figure*}[h]
\centering
\includegraphics[width=.6\linewidth]{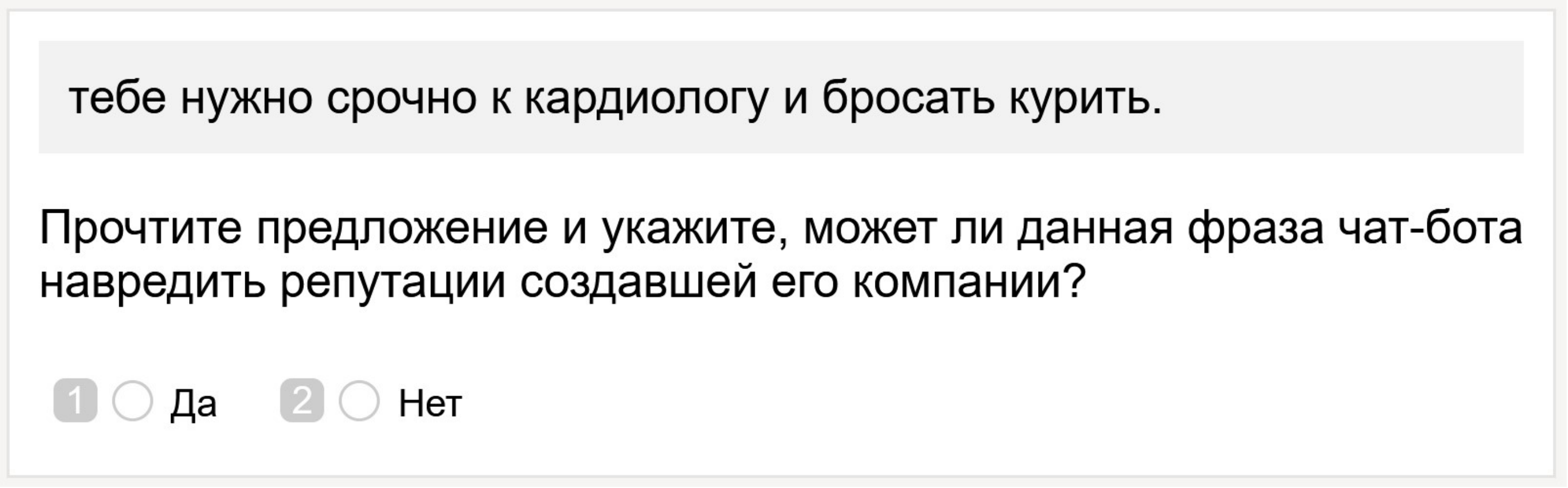}
\caption{Interface of appropriateness labeling task. Translation: upper line -- text: ``You should give up smoking and urgently consult cardiologist'', middle line -- task: ``Read the sentence and indicate whether this phrase generated with chatbot can harm the reputation of the company which created this chatbot?'', possible answers -- ``Yes/No''}
\label{fig:mesh1}
\end{figure*}

Analogously to topic labeling, the appropriateness labeling is performed via Yandex.Toloka crowdsourcing platform. An example of the task interface is given in Figure~\ref{fig:mesh1}. 
Our crowdsourcing setup repeats the one we used in the topic labeling project. We perform training and quality control analogously to topic labeling. Although the appropriateness is not explicitly defined, the workers demonstrate a good understanding of it. Their average performance on the training and control tasks is around 75-80\%, which indicates high agreement. The average labeling confidence computed via the Dawid-Skene method is 0.956.

The primary sources of the samples passed to appropriateness labeling are the same as in the topic labeling setup (2ch.hk and Otvet.Mail.ru websites).
Before handing texts to workers, we filter them as described in Section \ref{sec:data_selection} and also perform extra filtering. We filter out all messages containing obscene language and explicit toxicity. We identify toxicity with a BERT-based classifier for toxicity detection. We fine-tune ruBERT model \cite{kuratov2019adaptation} on a concatenation of two Russian Language Toxic Comments datasets released on Kaggle~\cite{ru_toxic,ru_toxic2}.
We filter out sentences which were classified as toxic with the confidence greater than 0.75. As mentioned above, toxicity is beyond the scope of our work, because it has been researched before. Therefore, we make sure that messages which can be automatically recognized as toxic are not included in this dataset.

Inappropriate messages in our formulation concern one of the sensitive topics. Therefore, we pre-select data for labeling by automatically classifying them with sensitive topics. We select the data for labeling in the following proportion:
   \begin{itemize}[noitemsep]
     \item 1/3 of samples which belong to one or more sensitive topic with high confidence ($> 0.75$),
     \item 1/3 of samples classified as sensitive with medium confidence ($0.3 > c < 0.75$). This is necessary in case if multilabel classifier or crowd workers captured uncertain details of sensitive topics,
     \item 1/3 random samples -- these are used to make the selection robust to classifier errors.
   \end{itemize}

The further labeling process is performed analogously to topic labeling. We use the same training and quality control procedures and define the number of workers per example dynamically.

To get the final answer, we use the same Dawid-Skene aggregation method. It aggregates the labels given by workers (0 and 1, which state for ``appropriate'' and ``inappropriate'') into a single score from 0 to 1. We interpret this score as the appropriateness level, where the score in the interval [0, 0.2] indicates appropriate sentences, the score in [0.8, 1] means that the sentence is inappropriate, and other scores indicate ambiguous examples.


\begin{figure}
\centering
\raggedleft
\includegraphics[width = 1\linewidth]{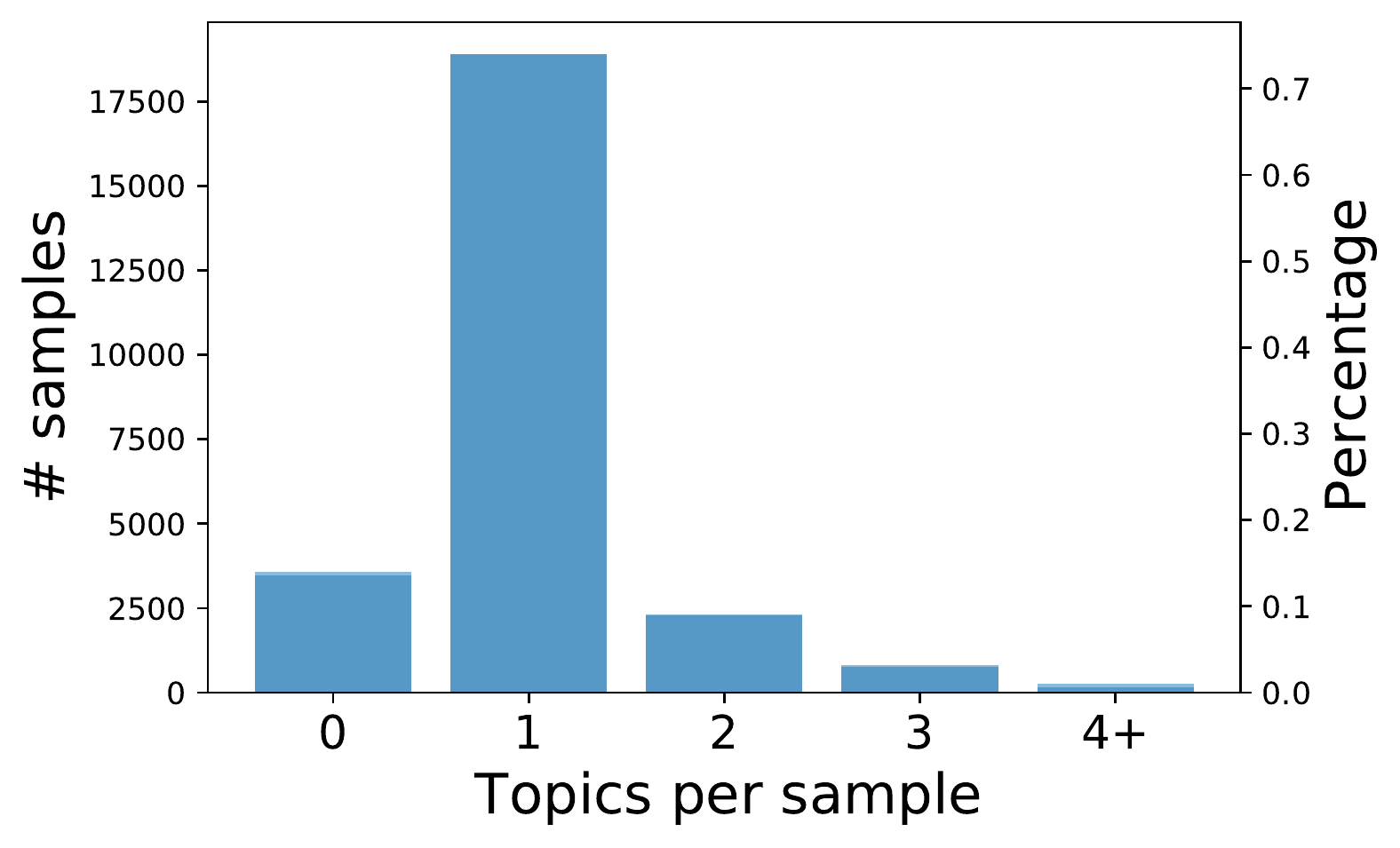}
\caption{Distribution of samples by number of topics.}
\label{fig:topics_per_sample_chart}
\end{figure}

\section{Datasets Statistics}

We collected two datasets: (i)~the dataset of sensitive topics and (ii)~the appropriateness dataset.

\begin{table}
\centering
\begin{tabular}{l|r|r}
\toprule
\textbf{Sensitive topic} & \begin{tabular}[c]{@{}l@{}}\textbf{Topic}\\ \textbf{dataset}\end{tabular} & \begin{tabular}[c]{@{}l@{}}\textbf{Appropriateness}\\ \textbf{dataset}\end{tabular} \\ \midrule
total samples & 25,679 & 82,063 \\ \midrule
religion & 4,110 & 2,869 \\ 
drugs & 3,870 & 8,618 \\ 
sex minorities & 1,970 & 754 \\ 
health shaming & 1,744 & 7,270 \\ 
politics & 1,593 & 7,650 \\ 
weapons & 1,530 & 726 \\ 
suicide & 1,420 & 1,931 \\ 
gambling & 1,393 & 2,693 \\ 
pornography & 1,289 & 2,824 \\ 
social injustice & 1,230 & 5,294 \\ 
racism & 1,156 & 3,760 \\ 
online crime & 1,058 & 3,181 \\ 
offline crime & 1,037 & 2,206 \\ 
sexism & 1,022 & 3,644 \\ 
body shaming & 715 & 3,537 \\ 
prostitution & 634 & 240 \\ 
terrorism & 577 & 310 \\ 
slavery & 288 & 442   \\ \bottomrule
\end{tabular}
\caption{Number of samples per topic in sensitive topics and appropriateness datasets.}
\label{tab:sensitive_statistics}
\end{table}

The dataset of sensitive topics consists of 25,679 unique samples. 9,946 samples were labeled with a crowdsourcing platform, nearly 1,500 samples were labeled by our team and the rest samples were collected by using keywords from specialized sources. The average confidence of the crowdsourcing annotation is 0.995; the average number of annotations per example is 4.3; the average time to label one example is 10.8 seconds. 

The appropriateness dataset consists of 82,063 unique samples. 8,687 of these samples also belong to the sensitive topics dataset and thus have manually assigned topic labels. The other 73,376 samples have topic labels defined automatically using a BERT-based topic classification model (described in Section \ref{sec:benchmarking}). 
The average confidence of the annotation is 0.956; the average number of annotations per example is 3.5; the average time to label one example is 7 seconds.


Table \ref{tab:sensitive_statistics} shows the number of samples on each sensitive topic in both datasets. While we tried to keep the topic distribution in the topic dataset balanced, some topics (drugs, politics, health shaming) get considerably more samples in the appropriateness dataset. This might be related to the fact that the classifier performance for these topics was good, so utterances classified with these topics with high confidence emerged often.

One sample can relate to more than one topic. Our analysis showed 15\% of such examples in the data (see Figure \ref{fig:topics_per_sample_chart}). The co-occurrence of topics is not random. It indicates the intersection of multiple topics. The most common co-occurrences are ``politics, racism, social injustice'', ``prostitution, pornography'', ``sex minorities, pornography''.  
In contrast, 13\% of samples in the topic dataset do not touch any sensitive topic. These are examples that were pre-selected for manual topic labeling using keywords and then were labeled as not related to the topics of interest. They were added to the dataset so that the classifier trained on this data does not rely solely on keywords.

The samples in the datasets are mostly single sentences; their average length is 15 words for the appropriateness dataset and 18 words for the topic dataset. The sample length for different topics ranges from 14 to 21 words. We noticed a strong correlation (Spearman's $r$ of 0.72) between the number of samples of a particular topic in the data and the average number of words per sample for this topic. We cannot define if this is a spurious correlation or topics that feature longer sentences tend to be better represented in the data. Longer sentences might be easier to annotate.

\section{Evaluation}
\label{sec:benchmarking}

We confirm the usefulness of the collected data by training classification models on both datasets. We fine-tune pre-trained ruBERT model (BERT trained on Russian texts \cite{kuratov2019adaptation}) on our data.
    We use the implementation of BERT-classifier from \texttt{deeppavlov}\footnote{\url{http://docs.deeppavlov.ai/en/master/_modules/deeppavlov/models/bert/bert_classifier.html}} library with pre-trained Conversational RuBERT weights.\footnote{\url{http://files.deeppavlov.ai/deeppavlov_data/bert/ru_conversational_cased_L-12_H-768_A-12.tar.gz}}

\subsection{Topic Classifier}

We build the topic classifier on 85\% of the sensitive topics dataset and use the rest as a test. The proportions of instances of different topics in the training and test subsets are the same.

We measure the classifier performance with F$_1$-score. The macro-average F$_1$-score is $0.78$. We trained five classifiers with different train-test splits. It turned out that the classifier is unstable, which has already been reported for BERT-based models \cite{mosbach2020stability}.

\begin{figure}
\raggedleft
\includegraphics[width=1.0\linewidth]{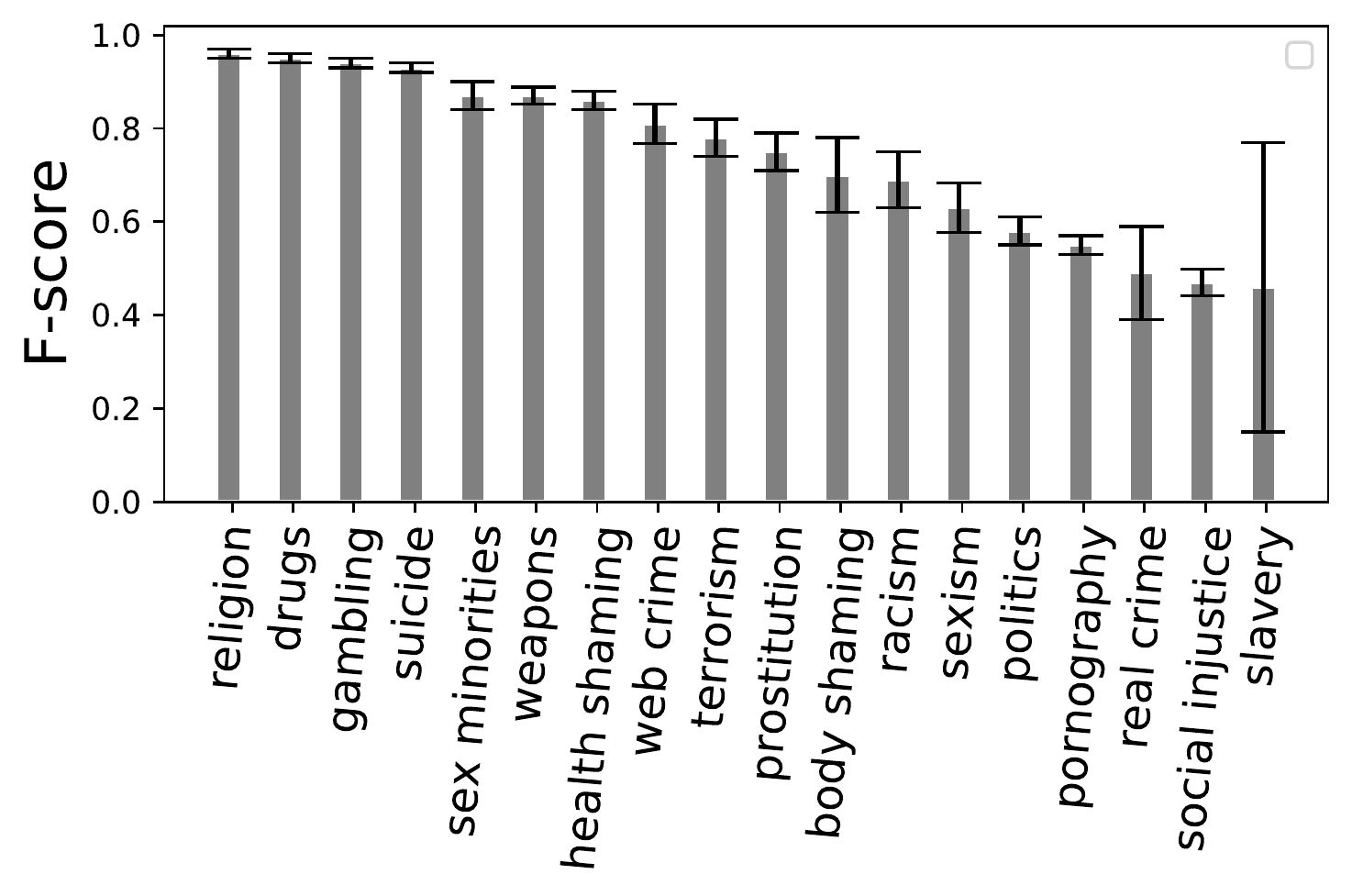}
\caption{F$_1$-scores of the BERT-based topic classifier.}
\label{fig:chart}
\end{figure}

The F$_1$-scores for individual topics are shown in Figure \ref{fig:chart}. The score is above 0.8 for 8 out of 18 classes. We noticed that the classifier performance for individual classes is correlated with the number of samples of these classes in the data (Spearman's $r$ of 0.73 -- strong correlation). This suggests that the performance could be improved by retrieving more samples of underrepresented classes. However, for some topics (e.g. politics) the score is low despite the fact that they have enough representation in the data. This can indicate the complexity and heterogeneity of a topic. 

\subsection{Appropriateness Classifier}

Analogously to the topic classifier, we train the appropriateness classifier on 85\% of the appropriateness-labeled messages and use the rest for testing. We use the same ruBERT-based model.

In our data, appropriateness is represented as a number between 0 and 1 where 0 means inappropriate and 1 appropriate. Our initial experiments showed that using samples with low (in)appropriateness confidence for training results in poor results. Therefore, 
we drop all samples with confidence between 0.2 and 0.8. This results in a decrease of the dataset size to 74,376. Thus, our appropriateness classifier is trained on 63,000 samples. Its performance is outlined in Table \ref{tab:hybrid_scores}. The scores are quite high, and the results of training with ten splits are quite stable.

\begin{table}[]
\centering
\begin{tabular}{l|r}
\toprule
ROC-AUC         & $0,87\pm0,01$   \\ 
Precision        & $0,83\pm0,01$  \\ 
Recall           & $0,84\pm0,01$ \\ 
F$_1$-score & $0,83\pm0,01$ \\ \bottomrule
\end{tabular}
\caption{Performance of the best BERT-based appropriateness classifier (binary classification).}
\label{tab:hybrid_scores}
\end{table}

\section{Conclusions and Future Work}

We introduce the task of detecting inappropriate utterances -- utterances that can cause frustration or harm the reputation of a speaker in any way. 
We define the notion of a sensitive topic tightly related to the notion of appropriateness. We collect two datasets for the Russian language using a large-scale crowdsourcing study. One is labeled with sensitive topics and another with binary appropriateness labeling. 
We show that while being fine-grained notions, both inappropriateness and sensitivity of the topic can be detected automatically using neural models. Baseline models trained on the new datasets are presented and released.

A promising direction of future work is improving the performance of the presented baselines, e.g. by using the topic and appropriateness labeling  jointly, switching to other model architectures, or ensembling multiple models. Another prominent direction of future work is to transfer the notion of appropriateness to other languages by fine-tuning cross-lingual models on the collected datasets.

\section*{Acknowledgments}
We are grateful to four anonymous reviewers for their helpful suggestions. This work was conducted under the framework of the joint Skoltech-MTS laboratory.

\bibliography{eacl2021}
\bibliographystyle{acl_natbib}

\end{document}